\documentclass[5p]{elsarticle}
\usepackage{lineno}
\usepackage{graphicx} 
\usepackage{subcaption}
\usepackage{ulem}
\usepackage{booktabs}
\usepackage{algorithm,algpseudocode}
\usepackage{amsmath}
\usepackage{tabularx}
\usepackage{multirow}
\usepackage{rotating}
\usepackage{pdflscape}
\usepackage{fancyhdr}
\usepackage{epstopdf}
\usepackage{breqn}

\modulolinenumbers[5]

\begin{document}
	
\captionsetup[table]{
	labelsep = newline,
	labelfont = bf,
	name = Table,
	justification=justified,
	singlelinecheck=false,
	skip = \medskipamount}	
	
	\newcolumntype{L}[1]{>{\raggedright\arraybackslash}p{#1}}
	\newcolumntype{C}[1]{>{\centering\arraybackslash}p{#1}}
	\newcolumntype{R}[1]{>{\raggedleft\arraybackslash}p{#1}}

\begin{frontmatter}
		
\title{Parallel Whale Optimization Algorithm for Solving Constrained and Unconstrained Optimization Problems}
		
\author[mymainaddress,mysecondaryaddress]{Amr M. Sauber}
\author[mymainaddress,mysecondaryaddress]{Mohammed M. Nasef}
\author[Add2,mysecondaryaddress]{Essam H. Houssein}
\author[Add3,mysecondaryaddress]{Aboul Ella Hassanien}
	
\address[mymainaddress]{Faculty of Science, Menoufia University, Egypt}
\address[Add2]{Faculty of Computers and Information, Minia University, Egypt}
\address[Add3]{Faculty of Computers and Information, Cairo University, Egypt}
\address[mysecondaryaddress]{Scientific Research Group in Egypt (SRGE) http://www.egyptscience.net}

\begin{abstract}
Recently the engineering optimization problems require large computational demands and long solution time even on high multi-processors computational devices. In this paper, an OpenMP inspired parallel version of the whale optimization algorithm (PWOA) to obtain enhanced computational throughput and global search capability is presented. It automatically detects the number of available processors and divides the workload among them to accomplish the effective utilization of the available resources. PWOA is applied on twenty unconstrained optimization functions on multiple dimensions and five constrained optimization engineering functions. The proposed parallelism PWOA algorithms performance is evaluated using parallel metrics such as speedup, efficiency. The comparison illustrates that the proposed PWOA algorithm has obtained the same results while exceeding the sequential version in performance. Furthermore, PWOA algorithm in the term of computational time and speed of parallel metric was achieved better results over the sequential processing compared to the standard WOA.
\end{abstract}

\end{frontmatter}

\section{Introduction}
\label{Sec:Introduction}

Optimization algorithms are needed everywhere and became a major part of nearly all applications. For instance; problems in basic science, engineering, medical sciences, and data science with all its countless applications. Upon using any optimization algorithm to solve problems, there are specific steps to be followed. The first step is to determine the objective function (or functions) of the problem and whether they should be maximized or minimized. The second step is to select a group of constraints that control the problem if they exist. The third step, determine the problem variables and investigate whether they are continues or discrete variables and the range of these variables. Optimization algorithms are generally utilized to obtain either the optimal or nearly optimal solution to a given problem as the exact solution are impossible or hard to find. There are many optimization algorithms. Researchers are still providing new algorithms every day hoping to find closer solutions. Meta-heuristic swarm-based algorithms recently emerged as a promising direction. This type of algorithms has proved to be highly efficient in approaching to the optimal solutions.

Meta-heuristic algorithms \cite{1} are divided into several categories. The first category is evolutionary algorithms, such as genetic algorithms (GA) \cite{1}, genetic programming (GP) \cite{2}, evolution strategy (ES) \cite{3}, probability-based incremental learning (PBIL) \cite{4}, and biogeography-based optimizer (BBO) \cite{5}. The second category is physics-based methods, this type mimic the physical rules in the universe. Some algorithms of this type are central force optimization (CFO) \cite{6}, gravitational search (GLSA) \cite{7}, gravitational algorithm (GSA) \cite{8}, and charged search (CSS) \cite{9}. The third category is human-based algorithms; this type depends on behavior of humans. This type contains some of the most popular algorithms like group search optimizer (GSO) \cite{10}, harmony search (HS) \cite{11}, and teaching-learning based optimization (TLBO) \cite{12}. The last category comprised of swarm-based algorithms that simulate the social behavior of animals. Particle swarm optimization (PSO) is considered the most popular algorithm in this category and originally, is developed by Kennedy and Eberhart \cite{13}, \cite{35}. Also, ant colony optimization \cite{14}, ant lion optimizer \cite{15}, dragonfly algorithm \cite{16}, moth-flame optimization algorithm \cite{17}, and finally whale optimization algorithm (WOA) \cite{18} in which in this paper we will propose a parallelization WOA version. 

WOA is one of promising swarm optimization algorithms. WOA simulates the behavior of the humpback whale. Hunting method is considered the most interesting characteristic for the humpback whales. Also, the bubble-net feeding method describes the foraging behavior \cite{19}. Humpback whales are similar to hunt krill or fishes schools closed to the surface. Furthermore, the foraging behaviors are accomplished by creating distinctive bubbles over a circle (9-shape). However, the foraging behavior has investigated by \cite{20}. Goldbogen et al. has captured 300 bubble-net nutrition of nine separate humpback whales utilizing tag sensors. the obtained results, two correlating maneuvers with bubble and named have been found called ‘upward-spirals’ and ‘double-loops’. Orderly former, humpback immersion dive around twelve m to down and the bubble in a circle is starting around the prey and in the end, it swims up toward the surface. The comprehensive information about the foraging behaviors is founded in \cite{20}. Mathematically, the spiral bubble-net feeding behavior has been modeled in order to solve many optimization problems. The location of prey has determined by Humpback whales and encircle them. In advance, the optimal position is not known in the search space and the present best agent solution has been assumed by WOA is considered the target prey or near to the optimal. whereas the best solution is obtained, other search agents aim to update their positions according to the best one.

Recently, the use of multi-processor architectures for general-purpose, high-performance parallel computing has been attracting researchers’ interest more and more, especially after the arrival datasets are becoming increasingly large, many applications and problems have huge amounts of data to be processed. So, with a bottleneck problem is being faced in terms of hardware capabilities for the above-mentioned algorithms. Also, the execution time is therefore often the only direct objective quantitative parameter on which comparisons can be based. The objective of this study is to design an OpenMP inspired parallel processing from the basic WOA to maximize its competitive advantage as it will be more efficient in handling big datasets. The proposed parallel algorithm is tested against twenty unconstrained functions and five constrained engineering problems \cite{21}. 

The structure of this paper is organized as follows; in Section \ref{Sec:BasicsandBackground} the basics and background are described. In Section \ref{Sec:PWOA}, the pseudo code for the parallel whale optimization algorithms is presented. In Section \ref{Sec:Results} the experimental results are illustrated to evaluate the proposed PWOA algorithm's performance. Finally, Section \ref{Sec:Conc} draws the conclusions and future work.

\section{Basics and Background}
\label{Sec:BasicsandBackground}
\subsection{OpenMp }
OpenMP is the most representative parallelization library of the SMP approach \cite{22}. 
OpenMP is also referred to as “fork-join parallelism” because a thread forks when a parallel region starts and threads join when a parallel region ends \cite{23}. OpenMP is composed of a set of directives \cite{24}, \cite{25}. Therefore, it is easy to program for parallelization using OpenMP. Also, OpenMP is scalable. The parallelized code using OpenMP can be applied to a single processor and multiple processors with the same source code. The use of single or multiple cores in calculations is determined depending on whether the OpenMP compiler is activated or not. In addition, OpenMP is portable, therefore, when a code is parallelized using OpenMP on one platform, the parallelized code can be used on other platforms with the use of the OpenMP compiler option turned on. In general, OpenMP is efficient for code parallelization because it does not re- quire programming the entire code. Instead, it can be developed just by modifying parts of the code. Thus, it requires less time to parallelize a code, and yields excellent performance with little effort \cite{26}. 

In the past, OpenMP had a limitation in the reduction of calculation time because the maximum number of core in the CPU was only four, six, or eight. However, the number of cores in CPU has increased dramatically as the relevant technology advances. Processors with dozens of core are now widely available. Currently, the interest in OpenMP is rising again because OpenMP is more convenient to program than MPI and requires only improvements to the original code rather than writing a new code. OpenMP has been successfully used to parallelize code. Recently, OpenMP was used to parallelize the finite element analysis program (FEAP) \cite{27}, and the authors compared the performance of the parallel code Warp3D and the parallelized FEAP code using OpenMP.

\subsection{Whale Optimization Algorithm (WOA)}
The mathematical model for WOA is illustrated in this section and for more detailed can found in \cite{19}. The following Equations describes the behavior of WOA to achieve encircling prey, spiral bubble-net feeding, and prey search.

	\begin{equation} 
	\label{EQ:eq1}
	\overrightarrow{D} = |\overrightarrow {C}. \overrightarrow{X^*}(t) -  \overrightarrow{X}(t) |
	\end{equation}
		
	\begin{equation}
	\label{EQ:eq2}
	\overrightarrow{X}(t+1) =  \overrightarrow{X^*}(t) - \overrightarrow {A}.  \overrightarrow{D}
	\end{equation}

	\begin{equation} 
	\label{EQ:eq3}
	\overrightarrow {A}= 2 \overrightarrow{a}. \overrightarrow{r} -\overrightarrow{a} 
	\end{equation}

	\begin{equation} 
	\label{EQ:eq4}
	\overrightarrow {C}= 2.\overrightarrow{r}  
	\end{equation}

	\begin{equation} 
	\label{EQ:eq5}
	\overrightarrow{X}(t+1) = \overrightarrow{D^'}.e^{b1} .cos(2\pi l)+ \overrightarrow{X^*}(t)
	\end{equation}

	\begin{equation} 
	\label{EQ:eq6}
	\overrightarrow{X}(t+1)=\left\{ \begin{matrix}
	\overrightarrow{{{X}^{*}}}(t)-\overrightarrow{A.}\overrightarrow{D}\,\,\,\,if\,P\prec 0.5  \\
	\overrightarrow{{{D}^{'}}}.{{e}^{b1}}.\cos (2\pi l)+\overrightarrow{{{X}^{*}}}(t)\,\,\,\,\,if\,P\ge 0.5  \\
	\end{matrix} \right.\,\
	\end{equation}

	Where\,$\overrightarrow{D^'}=\left| \overrightarrow{X^*}(t)-\overrightarrow{X}(t) \right| $

	\begin{equation} 
	\label{EQ:eq7}
	\overrightarrow{D}=\left| \overrightarrow{C.}\overrightarrow{{{X}_{rand}}}-\overrightarrow{X}(t) \right|\
	\end{equation}

	\begin{equation} 
	\label{EQ:eq8}
	\overrightarrow{X}(t+1)=\left| \overrightarrow{{{X}_{rand}}}-\overrightarrow{A}.\overrightarrow{D} \right|
	\end{equation}
	
	Where $\,P\in [0,\,1],\,l\in [-1,\,1],\,b=1.$

\subsection{Unconstrained Benchmark Functions}
Twenty optimization unconstrained functions are presented in this section. These unconstrained problems is a classical benchmark functions and utilizing in the optimization literature \cite{28}, \cite{29}. These unconstrained functions are described in Table \ref{TBL:Unimodal} and \ref{TBL:Multimodal} with the number of variable (N), variables ranges and the minimum value for each function.

\begin{table*}[!ht]
	\centering
	\caption{Unimodal unconstrained benchmark functions.}
	\label{TBL:Unimodal}
	\begin{tabular}{  p{.50cm} p{12cm} p{1.7cm} p{1cm}  p{1cm}}
		\toprule
		\textbf{No.} &  \textbf{Function} & \textbf{Range} & \textbf{N} & \textbf{Min (f)} \\ \midrule
		f1     & 
	
	${\sideset{}{}\sum_{i=1}^{n}} z_{j}^2$
	                & [-100,100]   & 30      & 0  \\\\
		f2   & $\sum\limits_{j=1}^{n}{\left| {{Z}_{j}} \right|}+\prod\limits_{j=1}^{n}{\left| {{Z}_{j}} \right|}$          & [-10,10]   & 30      &0     \\\\
		
		f3     & 	${\sideset{}{}\sum_{i=1}^{n}} (\sideset{}{}\sum_{j-1}^{i} Z_{j})^{2}$                   & [-100,100]   & 30       &0  \\\\
		
		f4     & $\underset{j}{max } \{ \left| {{Z}_{j}} \right|,1\le j\le n \}$                   & [-100,100]   & 30     & 0   \\\\
		
		f5     & 	${\sideset{}{}\sum_{j=1}^{n-1}} (100(Z_{j+1} - Z_{j}^{2})^{2} + (Z_j -1)^2) $                   & [-30,30]   & 30    &0 \\\\
		
		f6     & 	${\sideset{}{}\sum_{j=1}^{n}} ( Z_j + 0.5 )^2 $                   & [-100,100]   & 30     &0\\\\
		
		f7     & 	${\sideset{}{}\sum_{j=1}^{n}} jZ_j ^4 + rand[0,1) $                   & [-1.28,1.28]   & 30    &0 \\
		\bottomrule
	\end{tabular}
\end{table*}

\begin{table*}[!ht]
	\centering
	\caption{Multimodal unconstrained benchmark functions.}
	\label{TBL:Multimodal}
	\begin{tabular}{  p{.5cm} p{12cm} p{1.7cm} p{0.8cm}  p{1.5cm}}
		\toprule
		\textbf{No.} &  \textbf{Function} & \textbf{Range} & \textbf{N} & \textbf{Min (f)} \\ \midrule
		f8     &$\sum\nolimits_{j=1}^{n}{-{{Z}_{j}}}\sin (\sqrt{\left| {{Z}_{j}} \right|})$
		& [-500,500]&	30&	-418.98*5\\\\
		
		f9   & ${\sideset{}{}\sum_{i=1}^{n}} [ z_{j}^{2} - 10 cos(2 \pi z_j)+10 ]       $      &[-5.12,5.12]&	30&	0    \\\\

		f10     & $-20\exp (-0.2\sqrt{\frac{\sum\nolimits_{j=1}^{n}{z_{j}^{2}}}{n}})-\exp (\frac{1}{n}\sum\nolimits_{j=1}^{n}{\cos (2\pi {{Z}_{j}})+20+e}$                  & [-32,32]&	30	&0 \\\\
		
		f11     & $\frac{1}{4000}\sum\nolimits_{j=1}^{n}{z_{j}^{2}}-\prod\limits_{j=1}^{n}{\cos (\frac{{{z}_{j}}}{\sqrt{j}}})+1$                  & [-600,600]&	30	&0  \\\\
		
		f12     & $	\frac{\pi}{n} {{\{10\sin (\pi {{y}_{1}})+\sum\nolimits_{j=1}^{n-1}{({{y}_{i}}}-1)}^{2}}[1+10{{\sin }^{2}}(\pi {{y}_{j+1}})]+{{({{y}_{n}}-1)}^{2}}\}+\sum\nolimits_{j=1}^{n}{u({{x}_{j}},10,100,4)},  {{y}_{j}}=1+\frac{{{x}_{j}}+1}{4}$, $u({{x}_{j}},a,k,m)=\left\{ \begin{matrix}
		k{{({{x}_{j}}-a)}^{m}}\,\,\,{{x}_{j}}\succ a  \\
		0\,\,\,\,\,\,\,\,\,\,\,\,-a\prec {{x}_{j}}\prec a  \\
		k{{(-{{x}_{j}}-a)}^{m}}\,\,-a\succ {{x}_{j}}  \\
		\end{matrix} \right.$    &[-50,50]&	30&	0 \\ \\
		
		f13     & 	$0.1\{{{\sin }^{2}}(3\pi {{z}_{1}})+\sum\nolimits_{j=1}^{n}{{{({{z}_{j}}-1)}^{2}}}[1+{{\sin }^{2}}(3\pi {{z}_{j}}+1)]+{{({{z}_{n}}-1)}^{2}}[1+{{\sin }^{2}}(2\pi {{z}_{n}})]\}+\sum\nolimits_{j=1}^{n}{u({{z}_{j}},\,5,100,4)}$                   &[-50,50]&	30&	0\\\\
		
		f14     & 	$-cos(z_1) cos(z_2)exp(-(z_1-\pi)^2-(z_2-\pi)^2) $                   & [-100,100]&	2&	-1 \\\\
		
		f15     & 	$0.26(z_1^2+z_2^2)-0.48z_1 z_2 $                   & [-100,100]&	2&	0\\\\
		
		f16     & 	$4x_1^2-2.1x_1^4+1/3 x_1^6+x_1 x_2- 4x_2^2 + 4x_2^4 $                   &[-5,5]&	2&	−1 .0316 \\\\
		
		f17     & 	$(z_1+2z_2-7)^2+(2z_1+z_2-5)^2 $                   & [-10,10]&	2&	0 \\\\
		
		f18     & 	$0.5+\frac{{{\sin }^{2}}(\sqrt{{{a}^{2}}+{{b}^{2}}})-0.5}{{{(1+0.001({{a}^{2}}+{{b}^{2}}))}^{2}}}$               & [-100,100]&	2&	0\\\\

		f19     & 	$x_1^2+ 2x_2^2-0.3cos(3 \pi x_1)( 4 \pi x_2)+0.3 $                   &[-100,100]	&2	&0 \\\\
		
		f20     & 	$x_1^2+ 2x_2^2-0.3cos(3 \pi x_1+ 4 \pi x_2)+0.3 $                   &[-100,100]&	2&	0 \\
		\bottomrule
	\end{tabular}
\end{table*}

\subsection{Constrained Engineering Optimization Problems}
Five constrained engineering optimization problems are demonstrated in this section. 

\subsubsection{Gear Train Problem (Optim I)}
The main objective of this problem is to minimize The ratio of gear cost to train gear. The boundary constraints are the only parameters constructed and the decision variables are discrete for each gear have teeth integration. So, when the discrete variables have handled may the complexity is increased \cite{30}. This is the structural design problem in the literature formulated as follows: 

\begin{equation}
\label{EQ:eq9}
Min f(x) = ((\frac{1}{6.931}) - (\frac{x_{2}x_{3}}{x_{1}x_{4}}))^{2}
\end{equation}
Where, $12\le {{x}_{i}}\le 60$

\subsubsection{Cantilever Beam Design Problem (Optim II)}
Five hollow elements are included with square shaped in a cantilever beam problem. Each element is illustrated by one variable and a constant for the thickness, so five structural parameters are existing. Also, a vertical load utilized to a free end (node 6) of the beam and the right side (node 1) of the beam is rigidly defined \cite{31}. The mathematical structural design problem in the literature formulated as follows: 
\begin{dmath}
\label{EQ:eq10}
Minimize\,f({{x}_{1}},{{x}_{2}},{{x}_{3}},{{x}_{4}},{{x}_{5}})=0.6224({{x}_{1}}+{{x}_{2}}+{{x}_{3}}+{{x}_{4}}+{{x}_{5}})
\end{dmath}	
Subject to:\\ $g({{x}_{1}},{{x}_{2}},{{x}_{3}},{{x}_{4}},{{x}_{5}})=\frac{61}{x_{1}^{3}}+\frac{27}{x_{2}^{3}}+\frac{19}{x_{3}^{3}}+\frac{7}{x_{4}^{3}}+\frac{1}{x_{5}^{3}}-1\le 0,$\\
Where, $0.01\le {{x}_{1}},{{x}_{2}},{{x}_{3}},{{x}_{4}},{{x}_{5}}\le 100.$

\subsubsection{Himmelblau Problem (Optim III)}
This problem was first proposed by Himmelblau \cite{32}, the following Equation illustrates the minimize function for Himmelblau problem.
\begin{dmath}
\label{EQ:eq11}
f(X)=5.3578547x_{2}^{2}+0.835689{{x}_{1}}{{x}_{5}}+37.293239{{x}_{1}}-40729.141
\end{dmath}
Subject to:\\
$h1(X) = 85.334407 + 0.0056858 x_2 x_5 + 0.0006262 x_1 x_4 -0.0022053 x_3 x_5  $\\
$h2(X) = 80.51249 + 0.0071317 x_2 x_5  + 0.0029955 x_1 x_2  + 0.002181x_3^2 $\\
$h3(X) = 9.300961 + 0.0047026 x_3 x_5  + 0.0012547 x_1 x_3  + 0.0019085 x_3 x_4  $\\
Where, $0 \leq  h1(X) \leq 92 ,90 \leq h2(X) \leq 110,20 \leq h3(X) \leq 25$
$78 \leq x1 \leq 102; 33 \leq  x2 \leq  45, 27 \leq xi \leq 45; i = 3; 4; 5.$

\subsubsection{Constraint Problem (Optim IV)}
This problem is defined in \cite{33}.
\begin{dmath}
\label{EQ:eq12}
Min\,f(Z)={{({{z}_{1}}-10)}^{2}}+5{{({{z}_{2}}-12)}^{2}}+z_{3}^{4}+3{{({{z}_{4}}-11)}^{2}}
+10z_{5}^{6}+7z_{6}^{2}+z_{7}^{4}-4{{z}_{6}}{{z}_{7}}-10{{z}_{6}}-8{{z}_{7}}
\end{dmath}
Subject to:\\
$h1(Z) = 127 - 2z_1^2- 3z_2^4 - z_3 - 4z_4^2 - 5 z_5^  \ge 0$\\
$h2(Z) = 282 - 7z_1 - 3z_2 - 10z_3^2 - z_4 + z_5  \ge 0$\\
$h3(Z) = 196 - 23z_1 - z_2^2 -6z_6^2 + 8z_7  \ge 0$\\
$h4(Z) = -4z_1^2 - z_2^2 + 3z_1 z_2 - 2z_3^2 – 5z_6 + 11z_7  \ge 0$\\
Where, $- 10 \leq z_i \leq 10; i = 1; 2; 3; 4; 5; 6; 7$

\subsubsection{Constraint Problem (Optim V)}
This minimization problem consists of 729 disjoint spheres in which to construct the feasible region. This point $(z_1; z_2; z_3)$ called feasible point, if $r, q, and p$  have the inequality, \cite{34}.
the following Equation demonstrates the minimize function for this constrained optimization problem as follows:
\begin{equation}
\label{EQ:eq13}
Min\,f(Z)=(-100-{{({{z}_{1}}-5)}^{2}}-{{({{z}_{2}}-5)}^{2}}+{{({{z}_{3}}-5)}^{2}})/100
\end{equation}
Subject to:\\
$h(Z) = (z_{1}-p)^{2}+(z_{2}-q)^{2}+(z_{3}-r)^{2}-0.0625 \leq 0$\\
Where, $0 \leq z_i \leq 10; i = 1, 2, 3, and P, q, r = 1, 2, 3 \dots 9.$

\section{Parallel Whale Optimization Algorithm (PWOA)}
\label{Sec:PWOA}

\begin{figure}[ht!]
	\centering
	\includegraphics[width = 8cm, height =12.5cm]{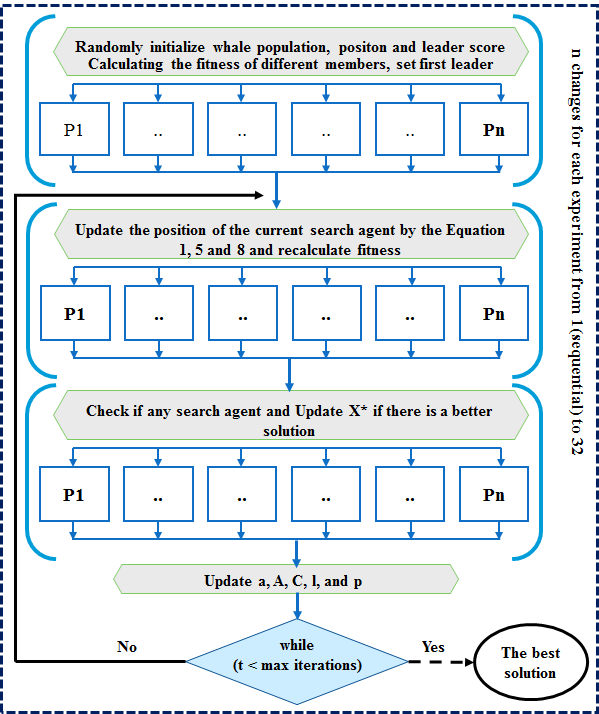}
	\caption{The flow diagram of PWOA.}
	\label{PWhale_Model}
\end{figure}

Parallel whale algorithm (PWOA) is inspired by OpenMP and designed to speed up the performance and efficiency of traditional whale optimization algorithm. The pseudo code for parallel whale algorithm is illustrated in Algorithm \ref{alg:PWOA}. The parallel WOA algorithm is represented by the flow diagram in Figure \ref{PWhale_Model}.

\begin{algorithm}[!htb]
	\caption{PWOA pseudo code.}
	\label{alg:PWOA}
	\begin{algorithmic}[1]
		\State Input: 
		\State Output: 
		\State Determine the $max$ number of processor available (NP).
		\State Initialize $X_i (i = 1, 2... n)$, the whales population.
		\State Figure the fitness utilizing NP Parallelization process.
		\While{$t<=< maximum number of iterations $}
		\State Utilizing NP Parallelization process.
		\For  { Search agent}
		\State Figure and adjust a; A, C, p and l.
		\If{$p<0.5$}
		\If{$(|A|< 1)$}
		\State Adjust the position by Eq 1. 
		\Else $(|A|\geq 1)$
		\State Select the random search agent ($X_{rand}$)
		\State Adjust the position by Eq 8. 
		\EndIf	
		\Else $(p \geq 0.5)$
		\State Adjust the position by Eq 5. 			
		\EndIf
		\EndFor
		\State Until all search agents adjusted.
		\State Figure the fitness for each agent.
		\State Adjust the best solution $(X*)$ if there exist a better one. 
		\State t=t+1
		\EndWhile	
	\end{algorithmic}
\end{algorithm}

\section{Experimental Results and Discussion}
\label{Sec:Results}
This section will introduce the experimental results to evaluate the performance of the proposed PWOA algorithm. For each function, the PWOA algorithm was run using the parameters depicted in Table \ref{TBL:Parameters.}.

\begin{table}[!htb]
	\centering
	\caption{Parameters setting of PWOA.}
	\label{TBL:Parameters.}
	\begin{tabular}{L{5cm}  L{2cm}}
		\toprule
		\textbf{Parameter} & \textbf{Value} \\ \midrule
		Number of Processors        & 1:32 \\
		Number of Runs        & 30 \\
		Number of Generations & 500 \\
		Number of Populations & 30 \\
		\bottomrule
	\end{tabular}
\end{table}

\subsection{Performance Measurements}
Three performance measurements, time, speed up, and efficiency are utilized to measures the performance of the proposed PWOA algorithm. 

\subsection*{Speedup:} Speedup measures the ration between the sequential execution time and the parallel execution time and obtained as follows:
\begin{equation}
\label{EQ:eq14}
S(P)=\ \frac{T(1)}{T(N)}
\end{equation}

Where, $T(1)$ is the execution time with one processor and $T(N)$ is the execution time with $N$ processors. 

\subsection*{Efficiency:} Efficiency is the ration between performance and the resources used to achieve that performance. Also it measures the usage of the computational resources and calculating by the following Equation.

\begin{equation}
\label{EQ:eq15}
E(N)=\ \frac{S(N)}{N}
\end{equation}
Where, $S(N)$ is the speedup for $N$ processors.

\subsection{Results and Discussion}
We evaluated the proposed novel version of the parallel WOA implementation on a ‘classical’ benchmark which comprised a set of
functions and optimization problems which are often used to evaluate stochastic optimization algorithms. 

Table \ref{Average8-20} and Table \ref{Average1-7},presents the comparison between the results for each unconstrained function on different number of processors. 

\begin{table*}[!ht]
	\centering
	\caption{the average results for each unconstrained unimodal benchmark functions on different processors.}
	\label{Average1-7}
	\begin{tabular}{p{3cm} p{2cm} p{2cm} p{2cm} p{2cm} p{2cm} p{2cm}}
		\toprule
		\textbf{Uni} &  \textbf{1 pr avg} & \textbf{2 pr avg} & \textbf{4 Pr avg} & \textbf{8 Pr avg} & \textbf{16 Pr avg} & \textbf{32 Pr avg} \\ \midrule
		F1     & 0&	0&	0&	0&	0&	0\\
		F2     & 0	&0.072074&	0.054917&	0.043621&	0.033433&	0.019102 \\
		F3     & 0.079938&	0.103134&	0.151403&	0.111385&	0.156628&	0.156449 \\
		F4     & 0.000185&	0.062434&	0.184934&	0.101786&	0.169903&	0.167528 \\
		F5     & 24.7332&	25.30602&	26.11116&	26.02233&	25.85528&	25.96711 \\
		F6     &0.002263&	0.127186&	0.196034&	0.195236&	0.213673&	0.178586\\
		F7     & 0.311751&	0.171542&	0.205471&	0.191338&	0.151258&	0.209223\\
		\bottomrule
	\end{tabular}
\end{table*}

\begin{table*}[!ht]
	\centering
	\caption{the the average results for each unconstrained multimodal benchmark functions on different processors.}
	\label{Average8-20}
	\begin{tabular}{  p{3cm} p{2cm} p{2cm} p{2cm}  p{2cm} p{2cm} p{2cm}}
		\toprule
		\textbf{Uni} &  \textbf{1 pr avg} & \textbf{2 pr avg} & \textbf{4 Pr avg} & \textbf{8 Pr avg} & \textbf{16 Pr avg} & \textbf{32 Pr avg} \\ \midrule
		
		F8     &  -12565.6&	-11116.9&	-9704.73&	-9688.25&	-10431.9&	-11110.2\\
		F9     &    0&	0.031745&	0.021057&	0.020036&	0.033443&	0.018585                    \\
		F10     &   2.81E-15&	0.063487&	0.041188&	0.031537&	0.016977&	0.036255                     \\
		F11     & -1.02131&	-0.93397&	-0.8071	&-0.87707&	-0.87715&	-0.85914                       \\
		F12     & 0.020837&	0.247476&	0.214033&	0.219846&	0.238623&	0.233929                       \\
		F13     & 0.000514&	0.051837&	0.032604&	0.022862&	0.031467&	0.074587                       \\	
		F14     & -1&	-1	&-1	&-1&	-1&	-1                       \\
		F15     &  0&	0&	0&	0&	0&	0                      \\
		F16     &   0&	0&	0&	0&	0&	0                     \\
		F17     &   2.01E-06&	2.09E-06&	1.11E-06&	9.57E-07&	3.28E-07&	4.95E-06                     \\
		F18     &      0&	0&	0&	0&	0&	0                     \\
		F19     &     0&	0&	0&	0&	0&	0                      \\
		F20     &       0&	0&	0&	0&	0&	0    \\					
		\bottomrule
	\end{tabular}
\end{table*}

Table \ref{Average3}, presents the optimization comparison results obtained for the constrained engineering optimization problems over different number of processors.

\begin{table*}[!ht]
	\centering
	\caption{the average result for each constrained optimization problem on different processors (P).}
	\label{Average3}
	\begin{tabular}{  p{3cm} p{2cm} p{2cm} p{2cm}  p{2cm} p{2cm} p{2cm}}
		\toprule
		\textbf{Uni} &  \textbf{1 pr avg} & \textbf{2 pr avg} & \textbf{4 Pr avg} & \textbf{8 Pr avg} & \textbf{16 Pr avg} & \textbf{32 Pr avg} \\ \midrule
		
		Gear Train     & 6.60E-21&	2.33E-19&	1.65E-22&	2.72E-20&	2.72E-20&	4.30E-20\\ \\
		
		Cantilever beam     & 13.03263&	13.03258&	13.03253&	13.03255&	13.03258&	13.03253 \\\\
		
		Himmelblau     & -30288.6	&-30288.6&	-30288.6&	-30288.6&	-30288.6&	-30288.6\\\\
	
		Constraint      & 707.1735&	707.458	&707.3283&	707.2103&	707.2301&	707.1876\\\\
		
		Constraint      & -1.16168&	-1.17116&	-1.17426&	-1.17431&	-1.17419&	-1.17426 \\
		\bottomrule
	\end{tabular}
\end{table*}


\begin{figure*}[!ht]
	\begin{subfigure} {0.5\textwidth}
		\includegraphics[width = 0.9\textwidth, height =3.6cm]{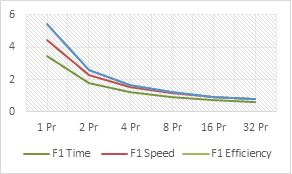}
		\caption{Function 1.}
		\label{F1}
	\end{subfigure}
	\begin{subfigure} {0.5\textwidth}
		\includegraphics[width = 0.9\textwidth, height =3.6cm]{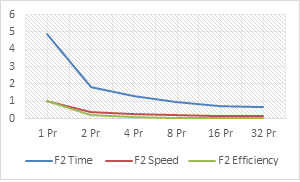}
		\caption{Function 2.}
		\label{F2}
	\end{subfigure}
	\caption{The graphical results obtained by the proposed PWOA algorithm in terms of time, speedup, and efficiency for different number of processors for each unconstrained multimodal benchmark functions from Function 1 to Function 2.}	
	\label{fig:Results1-2}	
\end{figure*}

\begin{figure*}[!ht]
	\begin{subfigure} {0.5\textwidth}
	\includegraphics[width = 0.9\textwidth, height =3.65cm]{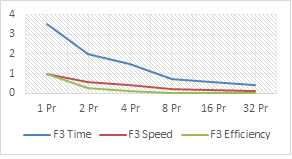}
		\caption{Function 3.}
		\label{F3}
	\end{subfigure}
	\begin{subfigure} {0.5\textwidth}
		\includegraphics[width = 0.9\textwidth, height =3.65cm]{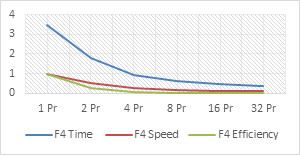}
		\caption{Function 4.}
		\label{F4}
	\end{subfigure}
	\begin{subfigure} {0.5\textwidth}
	\includegraphics[width = 0.9\textwidth, height =3.65cm]{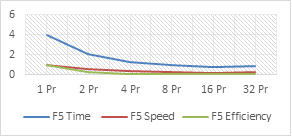}
		\caption{Function 5.}
		\label{F5}
	\end{subfigure}
	\begin{subfigure} {0.5\textwidth}
		\includegraphics[width = 0.9\textwidth, height =3.65cm]{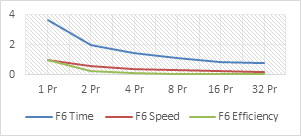}
		\caption{Function 6.}
		\label{F6}
	\end{subfigure}
	\begin{subfigure} {0.5\textwidth}
		\includegraphics[width = 0.9\textwidth, height =3.65cm]{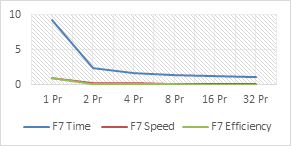}
		\caption{Function 7.}
		\label{F7}
	\end{subfigure}
	\caption{The graphical results obtained by the proposed PWOA algorithm in terms of time, speedup, and efficiency for different number of processors for each unconstrained multimodal benchmark functions from Function 3 to Function 7.}	
	\label{fig:Results3-7}	
\end{figure*}

Figure \ref{fig:Results1-2} and \ref{fig:Results3-7} shows the graphical results for the proposed PWOA algorithm in terms of time, speedup, and efficiency for different number of processors for each unconstrained unimodal benchmark functions. The execution time changing dependent on the problem dimension. This happens because of the limited internal resources can dedicate to each thread (processor). its clear from Figure \ref{fig:Results1-2} and \ref{fig:Results3-7}, the time, speed up and efficiency starts with high value associated with Pr1 and decreases proportionally when the number of processors increases.

Therefore, we have examined alternative solutions to solve the optimization problems. A parallel WOA is designed to deal with high-dimensional problems while allowing one to run each stage of a swarm (whale) in parallel. Even when one swarm (whale) is to be engaged, the full advantage of the parallel computation could be utilized, this an observation related to PWOA. Meanwhile, the parallelism cannot be utilized in computing the objective (fitness) function in this design. Whereas the fitness computation must be implemented as a sequential process associated to the swarm within the thread, subsequently, in such case of large-scale optimization problems PWOA becomes a very promising tool, according to the large amounts of computation and data.

\begin{figure*}[!ht]
	\begin{subfigure} {0.5\textwidth}
		\includegraphics[width = 0.9\textwidth, height =3.7cm]{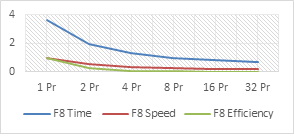}
		\caption{Function 8.}
		\label{F8}
	\end{subfigure}
	\begin{subfigure} {0.5\textwidth}
		\includegraphics[width = 0.9\textwidth, height =3.7cm]{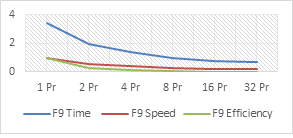}
		\caption{Function 9.}
		\label{F9}
	\end{subfigure}
	\begin{subfigure} {0.5\textwidth}
		\includegraphics[width = 0.9\textwidth, height =3.7cm]{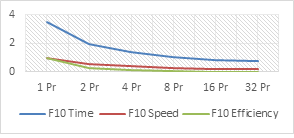}
		\caption{Function 10.}
		\label{F10}
	\end{subfigure}
	\begin{subfigure} {0.5\textwidth}
		\includegraphics[width = 0.9\textwidth, height =3.7cm]{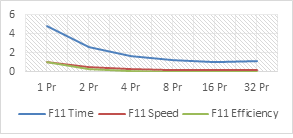}
		\caption{Function 11.}
		\label{F11}
	\end{subfigure}
	\begin{subfigure} {0.5\textwidth}
		\includegraphics[width = 0.9\textwidth, height =3.7cm]{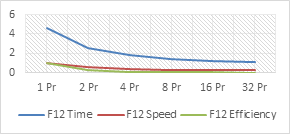}
		\caption{Function 12.}
		\label{F12}
	\end{subfigure}
	\begin{subfigure} {0.5\textwidth}
		\includegraphics[width = 0.9\textwidth, height =3.7cm]{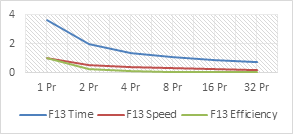}
		\caption{Function 13.}
		\label{F13}
	\end{subfigure}
	\begin{subfigure} {0.5\textwidth}
		\includegraphics[width = 0.9\textwidth, height =3.7cm]{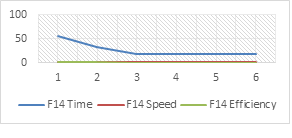}
		\caption{Function 14.}
		\label{F14}
	\end{subfigure}
	\begin{subfigure} {0.5\textwidth}
		\includegraphics[width = 0.9\textwidth, height =3.7cm]{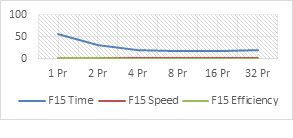}
		\caption{Function 15.}
		\label{F15}
	\end{subfigure}
	\begin{subfigure} {0.5\textwidth}
		\includegraphics[width = 0.9\textwidth, height =3.7cm]{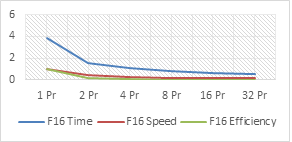}
		\caption{Function 16.}
		\label{F16}
	\end{subfigure}
	\begin{subfigure} {0.5\textwidth}
		\includegraphics[width = 0.9\textwidth, height =3.7cm]{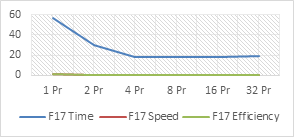}
		\caption{Function 17.}
		\label{F17}
	\end{subfigure}
	\caption{The graphical results obtained by the proposed PWOA algorithm in terms of time, speedup, and efficiency for different number of processors for each unconstrained multimodal benchmark functions from F8 to F17.}	
	\label{fig:Results8-17}	
\end{figure*}

\begin{figure*}[!ht]
	\begin{subfigure} {0.5\textwidth}
		\includegraphics[width = 0.9\textwidth, height =3.65cm]{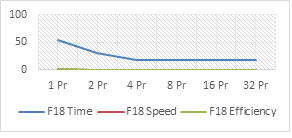}
		\caption{Function 18.}
		\label{F18}
	\end{subfigure}
	\begin{subfigure} {0.5\textwidth}
		\includegraphics[width = 0.9\textwidth, height =3.65cm]{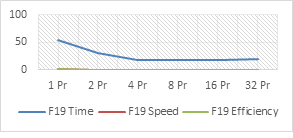}
		\caption{Function 19.}
		\label{F19}
	\end{subfigure}
	\begin{subfigure} {0.5\textwidth}
		\includegraphics[width = 0.9\textwidth, height =3.65cm]{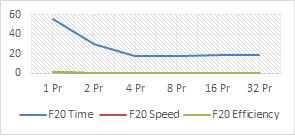}
		\caption{Function 20.}
		\label{F20}
	\end{subfigure}
	\caption{The graphical results obtained by the proposed PWOA algorithm in terms of time, speedup, and efficiency for different number of processors for each unconstrained multimodal benchmark functions from F18 to F20.}	
	\label{fig:Results18-20}	
\end{figure*}

Figure \ref{fig:Results8-17} and \ref{fig:Results18-20}shows the graphical results for the proposed PWOA algorithm in terms of time, speedup, and efficiency for different number of processors for each unconstrained multimodal benchmark functions.

\begin{figure*}[!ht]
	\begin{subfigure} {0.5\textwidth}
		\includegraphics[width = 0.9\textwidth, height =3.6cm]{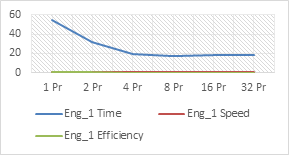}
		\caption{Optimization problem 1.}
		\label{Optim1}
	\end{subfigure}
	\begin{subfigure} {0.5\textwidth}
		\includegraphics[width = 0.9\textwidth, height =3.6cm]{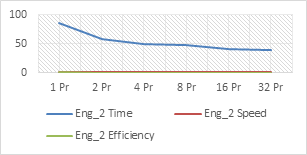}
		\caption{Optimization problem 2.}
		\label{Optim2}
	\end{subfigure}
	\begin{subfigure} {0.5\textwidth}
		\includegraphics[width = 0.9\textwidth, height =3.6cm]{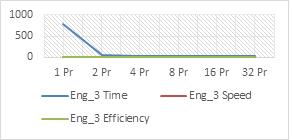}
		\caption{Optimization problem 3.}
		\label{Optim3}
	\end{subfigure}
	\begin{subfigure} {0.5\textwidth}
		\includegraphics[width = 0.9\textwidth, height =3.6cm]{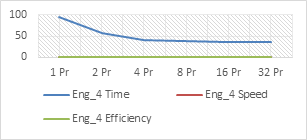}
		\caption{Optimization problem 4.}
		\label{Optim4}
	\end{subfigure}
	\begin{subfigure} {0.5\textwidth}
		\includegraphics[width = 0.9\textwidth, height =3.6cm]{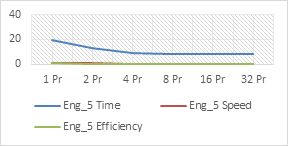}
		\caption{Optimization problem 5.}
		\label{Optim5}
	\end{subfigure}
	\caption{The graphical results for the proposed PWOA algorithm in terms of time, speedup, and efficiency for different number of processors for each constrained optimization problems.}	
	\label{fig:ResultsOptim1-5}	
\end{figure*}

Also, Figure \ref{fig:ResultsOptim1-5} shows the graphical results for the proposed PWOA algorithm in terms of time, speedup, and efficiency for different number of processors for each constrained optimization problems.

Actually, in most of the sequential evolutionary algorithms, the fitness considered as the significant task in which the performance of the algorithm depending on it as well as WOA. So, the number of fitness evaluations is the most primary measure of the execution time. In particular, we observed that the proposed PWOA, has achieved a better performance in terms of the time, speed up and efficiency when the number of processors increases.

\section{Conclusion}
\label{Sec:Conc}

Recently, the huge development for the parallel computational techniques and rapid increasing in the complex optimization problems in which are composed of performing many amounts of calculations. According to the aforementioned, in this paper, a novel parallel whale optimization algorithm (PWOA) is proposed, in order to enhance the WOA performance for solving optimization problems. The completed implementation was tested using 20 constrained benchmark functions and 5 unconstrained optimization problems. However, the time required to solve large-scale optimization problem was reduced extraordinarily by utilizing the parallel computation exhibiting multiple local minima. The experimental results have verified that PWOA has achieved a good performance in terms of the time, speed up and the efficiency. Furthermore, the proposed parallel algorithm based on OpenMP greatly improved the performance compared with the basic WOA and fully demonstrates the powerful calculating capacity of parallel processing. the proposed algorithm is an efficient algorithm for image processing, classification and great significance in improving the performance of intelligence systems dealing with big data and large-scale optimization problem. The future study will focus on hybrid recent swarm optimization algorithm in terms of the similar method presented in the paper in order to eliminate wasted CPU cycles via a dynamic task queue.

\end{document}